\documentclass[11pt]{article}
\usepackage{konvens2019}
\usepackage{mathptmx}
\usepackage[scaled=.90]{helvet}
\usepackage{courier}

\usepackage{url}
\usepackage{latexsym}
\usepackage{covington} 
\usepackage{booktabs} 
\usepackage{alltt} 
\usepackage{todonotes} 
\usepackage[utf8]{inputenc} 

\title{A Corpus for Automatic Readability Assessment and  Text Simplification of German}

\author{Alessia Battisti \\
  Institute of Computational Linguistics \\
  University of Zurich \\
  Andreasstrasse 15, 8050 Zurich \\
  {\tt alessia.battisti@uzh.ch} \\\And
  Sarah Ebling \\
  Institute of Computational Linguistics \\
  University of Zurich \\
  Andreasstrasse 15, 8050 Zurich \\
  {\tt ebling@cl.uzh.ch} \\}

\date{}

\begin{document}
\maketitle
\begin{abstract}
In this paper, we present a corpus for use in automatic readability assessment and automatic text simplification of German. The corpus is compiled from web sources and consists of approximately 211,000 sentences. As a novel contribution, it contains information on text structure, typography, and images, which can be exploited as part of machine learning approaches to readability assessment and text simplification. The focus of this publication is on representing such information as an extension to an existing corpus standard.
\end{abstract}

\section{Introduction}
\label{intro}
Simplified language is a variety of standard language characterized by reduced lexical and syntactic complexity, the addition of explanations for difficult concepts, and clearly structured layout.\footnote{The term \emph{plain language} is avoided, as it refers to a specific level of simplification. \emph{Simplified language}  subsumes all efforts of reducing the complexity of a piece of text.} Among the target groups of simplified language commonly mentioned are persons with cognitive impairment or learning disabilities, prelingually deaf persons, functionally illiterate persons, and foreign language learners \cite{bredel-maass-2016}. 

Two natural language processing tasks deal with the concept of simplified language: automatic readability assessment and automatic text simplification. Readability assessment refers to the process of  determining the level of difficulty of a text, e.g., along  readability measures, school grades, or levels of the Common European Framework of Reference for Languages (CEFR) \cite{cefr}. Readability  measures, in their traditional form, take into account only surface features. For example, the Flesch Reading Ease Score \cite{flesch-1948} measures the length of words (in syllables) and sentences (in words). While readability has been shown to correlate with such features to some extent \cite{just-carpenter-1980}, a consensus has emerged according to which they are not sufficient to account for all of the complexity inherent in a text. As \newcite[p.\,2618]{kauchak-et-al-2014}  state, ``the usability of readability formulas is limited and there is little evidence that the output of these tools directly results in improved understanding by readers''. Recently, more sophisticated models employing (deeper) linguistic features such as lexical, semantic, morphological, morphosyntactic, syntactic, pragmatic, discourse, psycholinguistic, and language model features have been proposed \cite{collins-thompson-2014,heimann-muehlenbock-2013,pitler-nenkova-2008,schwarm-ostendorf-2005,tanaka-et-al-2013}. 

Automatic text simplification was initiated in the late 1990s \cite{carroll-et-al-1998,chandrasekar-et-al-1996} and since then has been approached by means of rule-based and statistical methods. As part of a rule-based approach, the operations carried out typically include replacing complex lexical and syntactic units by simpler ones. A statistical approach generally conceptualizes the simplification task as one of converting a standard-language into a simplified-language text using machine translation. \newcite{nisioi-et-al-2017} introduced neural machine translation to automatic text simplification. Research on automatic text simplification is comparatively widespread for languages such as English, Swedish, Spanish, and Brazilian Portuguese. To the authors' knowledge, no productive system   exists for German. \newcite{suter-2015}, \newcite{suter-et-al-2016} presented a prototype of a rule-based  system for German. 

Machine learning approaches to both readability assessment and text simplification rely on data systematically prepared in the form of corpora. Specifically, for automatic text simplification via machine translation, pairs of standard-language/simplified-language texts aligned at the sentence level (i.e., parallel corpora) are needed. 

The paper at hand introduces a corpus developed for use in automatic readability assessment and automatic text simplification of German. The focus of this publication is on representing information that is valuable for these tasks but that hitherto has largely been ignored in machine learning approaches centering around simplified language, specifically, text structure (e.g., paragraphs, lines), typography (e.g., font type, font style), and image (content, position, and dimensions) information. The importance of considering such information has repeatedly been asserted theoretically \cite{arfe-et-al-2018,Bock2018,bredel-maass-2016}. 

The remainder of this paper is structured as follows: Section \ref{previous} presents previous corpora used for automatic readability assessment and  text simplification. Section \ref{our_work} describes our corpus, introducing its novel aspects and presenting the primary data (Section \ref{raw_text}), the metadata (Section \ref{metadata}), the secondary data (Section \ref{annotation}), the profile  (Section \ref{statistics}), and the results of machine learning experiments carried out on the corpus (Section \ref{empirical}).

\section{Previous Corpora for Automatic Readability Assessment and Automatic Text Simplification}
\label{previous}

	\begin{table*}
		\begin{tabular}{lrr}
			\toprule 
			& English & Simple English\\
			\midrule
			Number of sentences & 108,016 & 114,924\\
			Number of tokens & 2,645,771 & 2,175,240\\
			Avg. no. of words per sentence & 24.49 & 18.93\\
			Vocabulary size & 95,111 & 78,009\\
			\bottomrule
		\end{tabular}
		\caption{Parallel Wikipedia Simplification Corpus (PWKP) \cite{zhu-et-al-2010}: Profile (from \newcite{xu-et-al-2015})}
		\label{statistics_pwkp}
	\end{table*}

	\begin{table*}
		\begin{tabular}{lrrrrr}
			\toprule 
			& Original & Simple-1 & Simple-2 & Simple-3 & Simple-4\\
			\midrule
			Number of sentences & 56,037 & 57,940& 63,419& 64,035&64,162\\
			Number of tokens & 1,301,767& 1,126,148& 1,052,915& 903,417&764,103\\
			Avg. no. of sentences per document & 49.59& 51.27&56.12 & 56.67&56.78\\	
			Avg. no. of words per document & 1,152.01&996.59 & 931.78& 799.48&676.2 \\
			Avg. no. of words per sentence & 23.23& 19.44& 16.6& 14.11&11.91\\
			Vocabulary size & 39,046 & & & & 19,197 \\
			\bottomrule
		\end{tabular}
		\caption{Newsela Corpus \cite{xu-et-al-2015}: Profile}
		\label{statistics_newsela}
	\end{table*}

A number of corpora for  use in automatic readability assessment and automatic text simplification exist. The most well-known example is the Parallel Wikipedia Simplification Corpus (PWKP) compiled from parallel articles of the English Wikipedia and  Simple English Wikipedia \cite{zhu-et-al-2010} and consisting of around 108,000 sentence pairs. The  corpus profile is shown in Table~\ref{statistics_pwkp}. While the corpus represents the largest dataset involving simplified language to date, its application has  been criticized for various reasons \cite{amancio-specia-2014,xu-et-al-2015,stajner-et-al-2018}; among  these, the fact that Simple English Wikipedia articles are not necessarily direct translations of articles from the English Wikipedia stands out. \newcite{hwang-et-al-2015} provided an updated version of the corpus that includes a total of 280,000 full and partial matches between the two Wikipedia versions. Another frequently used data collection for English is the Newsela Corpus \cite{xu-et-al-2015} consisting of 1,130 news articles, each simplified into four school grade levels by professional editors. Table \ref{statistics_newsela} shows the profile of the Newsela Corpus. The table obviates that the difference in vocabulary size between the English and the simplified English side of the PWKP Corpus amounts to  only 18\%, while the corresponding number for the  English side and the level representing the highest amount of simplification in the Newsela Corpus (Simple-4) is 50.8\%. Vocabulary size as an indicator of lexical richness is generally taken to correlate positively with complexity \cite{vajjala-meurers-2012}.

\newcite{gasperin-et-al-2010} compiled the PorSimples Corpus consisting of Brazilian Portuguese texts (2,116  sentences), each with a natural and a strong simplification, resulting in around 4,500 aligned sentences. \newcite{drndarevic-saggion-2012}, \newcite{bott-et-al-2012}, \newcite{bott-saggion-2012} produced the Simplext Corpus consisting of 200 Spanish/simplified Spanish document pairs, amounting to a total of 1,149 (Spanish)/1,808 (simplified Spanish) sentences (approximately 1,000 aligned sentences).

\newcite{klaper-ebling-volk-2013} created the first parallel corpus for German/simplified German, consisting of 256 parallel texts downloaded from the web (approximately 70,000 tokens).

\section{Building a Corpus for Automatic Readability Assessment and Automatic Text Simplification of German}
\label{our_work}

Section \ref{previous} demonstrated that the only corpus containing simplified German available is that of \newcite{klaper-ebling-volk-2013}. Since its creation, a number of legal and political developments have spurred the availability of data in simplified German. Among these developments is the introduction  of a set of regulations for accessible information technology (\emph{Barrierefreie-Informationstechnik-Verordnung, BITV 2.0}) in Germany and the ratification of the United Nations Convention on the Rights of Persons with Disabilities (CRPD) in Switzerland. The paper at hand introduces a corpus that represents an  enhancement of the corpus of \newcite{klaper-ebling-volk-2013} in the following ways:
\begin{itemize}
	\item The  corpus contains  more parallel data.
	\item The  corpus  additionally contains  monolingual-only  data (simplified German).
	\item The corpus newly contains information on text structure, typography, and images.
\end{itemize}

The simplified German side of the parallel data together with the monolingual-only data can be used for automatic readability assessment. The parallel data in the corpus is  useful both for deriving rules for a rule-based  text simplification system in a data-driven manner and for training a data-driven machine translation system. A data augmentation technique such as back-translation \cite{sennrich-etal-2016-improving} can be applied to the monolingual-only data to arrive at additional (synthetic) parallel data.



\subsection{Primary Data}
\label{raw_text}

The corpus contains PDFs and webpages collected from web sources  in Germany, Austria, and Switzerland at the end of 2018/beginning of 2019. The web sources mostly consist of websites of governments, specialised institutions, translation agencies, and non-profit organisations (92 different domains). The documents cover a range of topics, such as politics (e.g., instructions for voting), health (e.g., what to do in case of pregnancy), and culture (e.g., introduction to art museums). 

For the webpages, a static dump of all documents was created. Following this, the documents were manually checked to verify the language. The main content   was subsequently extracted, i.e., HTML markup and boilerplate  removed using the  Beautiful Soup library for Python.\footnote{\url{https://pypi.org/project/beautifulsoup4/} (last accessed: February 27, 2019)} Information on text structure (e.g., paragraphs, lines) and typography (e.g., boldface, italics) was retained. Similarly, image information (content, position, and dimensions of an image) was preserved.

For PDFs, the  PDFlib Text and Image Extraction Toolkit (TET) was used to extract the plain text and record information on text structure, typography, and images.\footnote{\url{https://www.pdflib.com/} (last accessed: February 27, 2019)} The toolkit produces output in an XML format (TETML). 

\subsection{Metadata}
\label{metadata}
\begin{figure*}
\begin{alltt}
<?xml version='1.0' encoding='utf-8'?>
<olac:olac xmlns:cld="http://purl.org/cld/terms/"
        xmlns:dc="http://purl.org/dc/elements/1.1/"
        xmlns:dcterms="http://purl.org/dc/terms/"
        xmlns:oai="http://www.openarchives.org/OAI/2.0/"
        xmlns:oai_dc="http://www.openarchives.org/OAI/2.0/oai_dc/"
        xmlns:olac="http://www.language-archives.org/OLAC/1.1/"
        xmlns:schemaLocation="http://www.language-archives.org/OLAC/
        1.1/olac.xsd"
        xmlns:tei="http://www.tei-c.org/ns/1.0"
        xmlns:xs="http://www.w3.org/2001/XMLSchema"
        xmlns:xsi="http://www.w3.org/2001/XMLSchema-instance"
        xmlns="http://www.openarchives.org/OAI/2.0/static-repository">
    <dc:title xml:lang="de">Maria sagt es weiter...
        Ein Bilder-Lese-Buch über sexuelle Gewalt und Hilfe holen.
    </dc:title>
    <dc:language xsi:type="olac:language"
        olac:code="de">A2</dc:language>
    <dc:publisher>Frauenbüro der Stadt Linz</dc:publisher>
    <dc:publisher xsi:type="dcterms:URI">www.linz.at/frauen</dc:publisher>
    <dc:contributor xsi:type="olac:role" olac:code="author">
        Verein Hazissa</dc:contributor>
    <dc:contributor xsi:type="olac:role" olac:code="translator">
        capito Oberösterreich</dc:contributor>
    <dc:contributor xsi:type="olac:role" olac:code="illustrator">
        Müller, Silke</dc:contributor>
    <dc:identifier xsi:type="dcterms:URI">
    https://www.linz.at/images/MariaD.pdf</dc:identifier>
    <dc:date xsi:type="dcterms:W3CDTF">2016</dc:date>
    <dc:format xsi:type="dcterms:IMT">application/pdf</dc:format>
    <dc:type xsi:type="dcterms:DCMIType">Text</dc:type>
    <dc:type xsi:type="dcterms:DCMIType">StillImage</dc:type>
    <dc:type xsi:type="olac:linguistic-type" olac:code="primary_text"/>
    <dc:source>mariad.tetml</dc:source>
    <dc:rights/>
    <dcterms:tableOfContents>
        Maria sagt es weiter Seite 7; Informationen zu sexueller Gewalt 
        Seite 12; Adressen von Beratungs-Stellen Seite 17; Wörterbuch 
        Seite 32
    </dcterms:tableOfContents>
</olac:olac>
    \end{alltt}
    \caption{Sample metadata in OLAC for a PDF document from the corpus}
	\label{sample_metadata}
\end{figure*}

Metadata was collected automatically from the HTML (webpages) and TETML (PDFs)  files, complemented manually, and recorded in the Open Language Archives Community (OLAC) Standard.\footnote{\url{http://www.language-archives.org/OLAC/olacms.html} (last accessed: February 28, 2019)} OLAC is based on a reduced version of the Dublin Core Metadata Element Set (DCMES).\footnote{\url{http://dublincore.org/} (last accessed: February 28, 2019)} Of the 15 elements of this ``Simple Dublin Core'' set, the following 12 were actively used along with controlled vocabularies of OLAC and Dublin Core:
\begin{itemize}
	\item \texttt{title}: title of the document, with the language  specified as the value of an \texttt{xml:lang} attribute and alternatives to the original title (e.g., translations) stored as \texttt{dcterms:alternative} (cf.\,Figure \ref{sample_metadata} for an example)
	\item \texttt{contributor}: all person entities linked to the creation of a document, with an \texttt{olac:code} attribute with values from the OLAC role vocabulary used to further specify the role of the contributor, e.g., \texttt{author}, \texttt{editor}, \texttt{publisher}, or \texttt{translator}
	\item \texttt{date}: date mentioned in the metadata of the HTML or PDF source or, for news and blog articles,  date mentioned in the body of the text, in W3C date and time format 
	\item \texttt{description}: value of the description in the metadata of an HTML document or list of sections of a PDF document, using the Dublin Core qualifier \texttt{TableOfContents}
	\item \texttt{format}: distinction between the Internet Media Types (MIME types) \texttt{text/html} (for webpages) and \texttt{application/pdf} (for PDFs)
	\item \texttt{identifier}: URL of the document or International Standard Book Number (ISBN) for books or brochures
	\item \texttt{language}: language of the document as value of the attribute \texttt{olac:code} (i.e., \texttt{de}, as conforming to ISO 639), with the CEFR level as optional element content
	\item \texttt{publisher}: organization or person that  made the  document available
	\item \texttt{relation}: used to establish a link between documents in German and simplified German for the parallel part of the corpus, using the Dublin Core qualifiers \texttt{hasVersion} (for the German text) and \texttt{isVersionOf} (for the simplified German text)
	\item \texttt{rights}: any piece of information about the rights of a document, as far as available in the source 
	\item \texttt{source}: source document, i.e., HTML for web documents and TETML for PDFs
	\item \texttt{type}: nature or genre of the content of the document, which, in accordance with the DCMI Type Vocabulary, is \texttt{Text} in all cases and additionally \texttt{StillImage} in cases where a document also contains images. Additionally, the linguistic type is specified according to the OLAC Linguistic Data Type Vocabulary, as either \texttt{primary\_text} (applies to most documents) or \texttt{lexicon} in cases where a document represents an entry of a simplified language vocabulary 
\end{itemize}

The  elements \texttt{coverage} (to denote the spatial or temporal scope of the content of a resource), \texttt{creator} (to denote the author of a text, see \texttt{contributor} above), and \texttt{subject} (to denote the topic of the document content)  were not used.

Figure \ref{sample_metadata} shows an example of OLAC metadata. The source document described with this metadata record is a PDF structured into chapters, with text corresponding to the CEFR level A2 and images. Metadata in  OLAC can be converted into the metadata standard of CLARIN (a European research infrastructure for language resources and technology),\footnote{\url{https://www.clarin.eu/} (last accessed: February 27, 2019)} the  Component MetaData Infrastructure (CMDI).\footnote{\url{https://www.clarin.eu/faq/how-can-i-convert-my-dc-or-olac-records-cmdi} (last accessed: February 28, 2019)} The CMDI standard was chosen since it is the supported metadata version of CLARIN, which is specifically popular in German-speaking countries. 

Information on the language level of a simplified German text (typically A1, A2, or B1) is particularly valuable, as it allows for conducting automatic readability assessment and graded automatic text simplification experiments on the data. 52 websites and 233 PDFs (amounting to approximately 26,000 sentences) have an explicit language level label.


\subsection{Secondary Data}
\label{annotation}

Annotations were added in the Text Corpus Format by WebLicht (TCF)\footnote{\url{https://weblicht.sfs.uni-tuebingen.de/weblichtwiki/index.php/The_TCF_Format} (last accessed: April 11, 2019)} developed as part of CLARIN. TCF supports standoff annotation, which allows for representation of annotations with conflicting hierarchies.  TCF does not assign a separate file for each annotation layer; instead, the source text and all annotation layers are stored jointly in a single file. A token layer acts as the  key element to which all other annotation layers are linked. 

The following types of annotations were added: text structure, fonts, images, tokens, parts of speech, morphological units, lemmas, sentences, and dependency parses. TCF does not readily accommodate the incorporation of all of these types of information. We therefore extended the format in the following ways:
\begin{itemize}
	\item Information on the font type and font style (e.g., italics, bold print) of a token and its position on the physical page (for PDFs only) was specified as attributes to the \texttt{token} elements of the \texttt{tokens} layer (cf.\,Figure \ref{sample_annotation} for an example)
	\item Information on physical page segmentation (for PDFs only), paragraph segmentation, and line segmentation was added as part of a \texttt{textspan} element in the \texttt{textstructure} layer 
	\item A separate \texttt{images} layer was introduced to hold \texttt{image} elements that take as attributes the x and y coordinates of the images, their dimensions (width and height), and the number of the page on which they occur 
	\item A separate \texttt{fonts} layer was introduced to preserve detailed information on the font configurations referenced in the \texttt{tokens} layer
\end{itemize}

\begin{figure*}
\begin{alltt}
<TextCorpus>
    <text>...</text>
    <tokens>
        <token ID="t_0" font="F0">Vorwort</token>
        <token ID="t_1" font="F0">Liebe</token>
        <token ID="t_2" font="F0">Leserinnen</token>
        ...
    </tokens>
    <sentences>
        <sentence ID="s_0" tokenIDs="t_0 t_1 t_2 t_3"/>
        ...
        </sentences>
    <textstructure>
        <textspan start="t_0" type="paragraph" end="t_0"/>
        <textspan start="t_0" type="line" end="t_0"/>
        <textspan type="paragraph" start="t_1" end="t_3"/>
        <textspan type="line" start="t_1" end="t_3"/>
        <textspan type="paragraph" start="t_4" end="t_27"/>
        ...
    </textstructure>
    <lemmas>
        <lemma ID="l_0" tokenIDs="t_0">Vorwort</lemma>
        <lemma ID="l_1" tokenIDs="t_1">lieb</lemma>
        <lemma ID="l_2" tokenIDs="t_2">Leserin</lemma>
        ...
    </lemmas>
    <POStags tagset="stts">
        <tag tokenIDs="t_0">NN</tag>
        <tag tokenIDs="t_1">ADJA</tag>
        <tag tokenIDs="t_2">NN</tag>
        ...
    </POStags>
    <morphology>
        <analysis tokenIDs="t_0">Neut|_|Sg</analysis>
        <analysis tokenIDs="t_1">Pos|Fem|_|Pl|St|</analysis>
        ...
    </morphology>
    <depparsing>
        ...
    </depparsing>
    <images>
        <image ID="I0" page="1" x="-1.07" y="112.47"/>
        ...
    </images>
    <fonts>
        <font id="F0" name="TradeGothic-BoldTwo"
        fullname="UDSPGZ+TradeGothic-BoldTwo" type="Type 1 CFF"
        embedded="true" ascender="977" capheight="722" italicangle="0"
        descender="-229" weight="700" xheight="520"/>
        ...
    </fonts>
</TextCorpus>
	\end{alltt}
	\caption{Sample corpus annotation}
	\label{sample_annotation}
\end{figure*}

Linguistic annotation was added automatically using the ParZu dependency parser for German \cite{sennrich-2009b} (for tokens and dependency parses), the NLTK toolkit \cite{bird-et-al-2009} (for sentences), the TreeTagger \cite{schmid-1995} (for part-of-speech tags and lemmas), and Zmorge \cite{sennrich-kunz-2014} (for morphological units). Figure \ref{sample_annotation} shows a sample corpus annotation. Together, the metadata shown in Figure \ref{sample_metadata} and the annotations presented in Figure \ref{sample_annotation} constitute a complete TCF file.

\subsection{Corpus Profile}
\label{statistics}
	\begin{table*}
\centering
		\begin{tabular}{lrr}
			\toprule 
			& German & Simplified German\\
			\midrule
			\multicolumn{3}{l}{Monolingual}\\
			\midrule
			Number of documents & & 5,461\\
			Number of sentences &  & 172,773\\
			Number of tokens &  & 1,916,045\\
			Avg. no. of sentences per document & & 31.64\\	
			Avg. no. of tokens per sentence & &11.09 \\
			\midrule			
			\multicolumn{3}{l}{Parallel}\\
			\midrule
			Number of documents & 378 & 378 \\
			Number of sentences &  17,121& 21,072\\
			Number of tokens & 347,941 & 246,405\\
			Avg. no. of sentences per document & 45.29& 55.75\\	
			Avg. no. of tokens per sentence & 20.32 & 11.69\\
			Vocabulary size & 33,384 & 16,352\\
			\midrule
			\multicolumn{3}{l}{Parallel (total)}\\
			\midrule
			Number of documents & \multicolumn{2}{r}{756} \\
			Number of sentences & \multicolumn{2}{r}{38,193}  \\
			Number of tokens &  \multicolumn{2}{r}{594,346} \\
			Avg. no. of sentences per document & \multicolumn{2}{r}{50.52} \\	
			Avg. no. of tokens per sentence & \multicolumn{2}{r}{15.56} \\
			\midrule
			\multicolumn{3}{l}{Monolingual and parallel (total)}\\
			\midrule
			Number of documents &  \multicolumn{2}{r}{6,217} \\
			Number of sentences &  \multicolumn{2}{r}{210,966} \\
			Number of tokens &   \multicolumn{2}{r}{2,510,391} \\
			Avg. no. of sentences per document &  \multicolumn{2}{r}{33.93} \\	
			Avg. no. of tokens per sentence &  \multicolumn{2}{r}{11.90} \\
			\bottomrule
		\end{tabular}
		\caption{Corpus profile}
		\label{statistics_our_corpus}
	\end{table*}

The resulting corpus contains 6,217 documents (5,461 monolingual documents plus 378 documents for each side of the parallel data). Table \ref{statistics_our_corpus} shows the  corpus profile. The monolingual-only documents on average contain fewer sentences than the simplified German side of the parallel data (average document length in sentences 31.64 vs.\ 55.75). The average sentence length is almost equal (approx.\ 11 tokens). Hence, the monolingual-only texts are shorter than the simplified German texts in the parallel data. Compared to their German counterparts, the simplified German texts in the parallel data have clearly undergone a process of lexical simplification: The vocabulary is smaller by 51\%  (33,384 vs. 16,352 types), which is comparable to the rate of reduction reported in Section \ref{previous} for the Newsela Corpus (50.8\%).

\subsection{Empirical validation of the corpus}
\label{empirical}
\newcite{battisti-2019} applied unsupervised machine learning techniques to  the simplified German texts of the corpus presented in this paper  with the aim of investigating evidence of multiple complexity levels. While the detailed results are beyond the scope of this paper, the author found features based on the structural information that is a unique property of this corpus (e.g., number of images, number of paragraphs, number of lines, number of words of a specific font type, and adherence to a one-sentence-per-line rule) to be predictive of the level of difficulty of a simplified German text. To our knowledge, this is the first study to deliver empirical proof of the relevance of such features.

\section{Conclusion and Outlook}
We have introduced a corpus compiled for use in automatic readability assessment and automatic text simplification of German. While such tasks have been addressed for other languages, research on German is still scarce. The features exploited as part of machine learning approaches to readability assessment so far typically include surface and/or (deeper) linguistic features. The corpus presented in this paper additionally contains information on text structure, typography, and images. These features have been shown to be  indicative of simple vs. complex texts both theoretically and, using the corpus described in this paper, empirically.

Information on text structure, typography, and images can also be leveraged as part of a neural machine translation approach to text simplification. A set of parallel documents used in machine translation additionally requires sentence alignments, which are still missing from our corpus. Hence, as a next step, we will include such information using the Customized Alignment for Text Simplification (CATS) tool \cite{stajner-et-al-2017}.


\bibliographystyle{konvens2019}
\bibliography{konvens2019.bib}

\end{document}